\documentclass[10pt,twocolumn,letterpaper]{article}

\usepackage{cvpr}              % To produce the CAMERA-READY version
\usepackage{multirow}
\usepackage{adjustbox} 
\usepackage{algorithm}  
\usepackage{algorithmic}

\definecolor{cvprblue}{rgb}{0.21,0.49,0.74}
\usepackage[pagebackref,breaklinks,colorlinks,allcolors=cvprblue]{hyperref}

\def\paperID{7369} % *** Enter the Paper ID here
\def\confName{CVPR}
\def\confYear{2025}

\title{Reducing Class-wise Confusion for Incremental Learning \\
with Disentangled Manifolds}

\author{
Huitong Chen, Yu Wang\thanks{Corresponding author}, Yan Fan, Guosong Jiang, Qinghua Hu\\
Tianjin Key Lab of Machine Learning, College of Intelligence and Computing, Tianjin University, China\\
Haihe Laboratory of Information Technology Application Innovation (Haihe Lab of ITAI), Tianjin, China \\
{\tt\small \{chtcs, wang.yu, fyan\_0411, jiangggss, huqinghua\}@tju.edu.cn}  
}  

\begin{document}
\maketitle
% compile the main paper
\begin{abstract}
Class incremental learning (CIL) aims to enable models to continuously learn new classes without catastrophically forgetting old ones. A promising direction is to learn and use prototypes of classes during incremental updates. Despite simplicity and intuition, we find that such methods suffer from inadequate representation capability and unsatisfied feature overlap. These two factors cause class-wise confusion and limited performance. In this paper, we develop a Confusion-REduced AuTo-Encoder classifier (CREATE) for CIL. Specifically, our method employs a lightweight auto-encoder module to learn compact manifold for each class in the latent subspace, constraining samples to be well reconstructed only on the semantically correct auto-encoder. Thus, the representation stability and capability of class distributions are enhanced, alleviating the potential class-wise confusion problem. To further distinguish the overlapped features, we propose a confusion-aware latent space separation loss that ensures samples are closely distributed in their corresponding low-dimensional manifold while keeping away from the distributions of features from other classes. Our method demonstrates stronger representational capacity and discrimination ability by learning disentangled manifolds and reduces class confusion. Extensive experiments on multiple datasets and settings show that CREATE outperforms other state-of-the-art methods up to $5.41\%$. The code is available at \href{https://github.com/lilyht/CREATE}{https://github.com/lilyht/CREATE}.
\end{abstract}    
\section{Introduction}
\label{sec:intro}

Class Incremental Learning (CIL) aims to enable deep learning models to continuously learn new classes while maintaining old knowledge. It has significant implications in intelligent systems that require continuous evolution. For example, in an autonomous driving scenario, the system should gradually adapt to new environments, infrastructures, and traffic patterns in different countries without forgetting its previous driving capabilities. 
A fundamental challenge in CIL is to tackle catastrophic forgetting \citep{french1999catastrophic, kirkpatrick2017overcoming, lin2024class}, where the performance of previously learned knowledge significantly deteriorates when the model adapts to new class instances. 

\begin{figure*}[t]
\begin{center}
\includegraphics[width=1.0\linewidth]{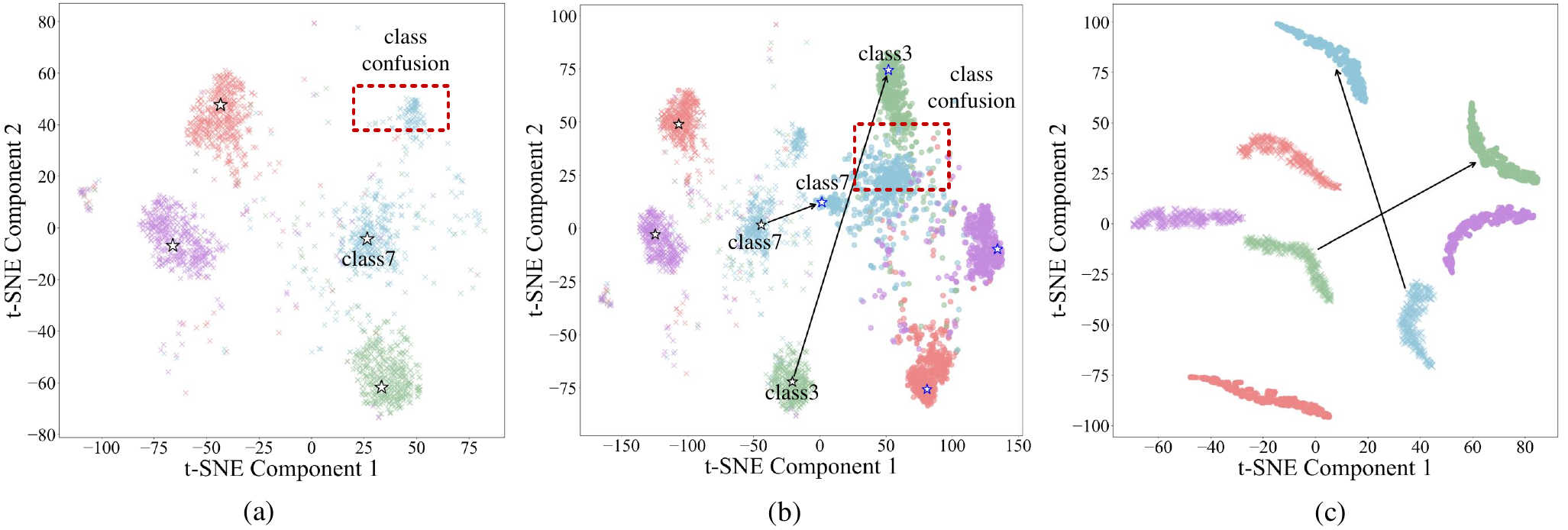}
\end{center}
\caption{
T-SNE visualization of feature distributions under CIFAR100 Base10 Inc10. The points marked with crosses represent features of the initial phase, while the points marked with circles indicate features of the final phase. Samples within the red box represent two instances of class confusion.
(a) Class distributions and prototypes in the initial phase. Using a prototype to represent a class is incomplete.
(b) A drift in incremental learning leads to alterations in class distribution and feature overlap. 
(c) Our method exhibits the ability to disentangle manifolds and reduce confusion. 
}
\label{fig:1-intro}
\end{figure*}

Existing studies are dedicated to mitigating this problem and primarily address the issue from three perspectives. 
Specifically, \textbf{knowledge retention-based approaches} \citep{chen2022multi,gao2023exploring,sarfraz2023error,zhu2024reshaping, wen2024class,fan2024dynamic, fan2024persistence} aim to discover and maintain inherent knowledge structures through regularization, thereby reducing changes in the model's intrinsic knowledge structure and alleviating the forgetting of old knowledge. Such methods impose significant constraints on the model's old knowledge, creating challenges in knowledge refinement when new knowledge is introduced. Therefore, \textbf{model expansion-based approaches }\citep{yan2021der, wang2022foster, wang2023beef} are designed to dynamically adjust the representational capacity of a model to adapt to new tasks. However, this type of approach typically involves a large number of parameters, and there is redundancy between the newly expanded and old branches.
\textbf{Prototype-based approaches} \citep{rebuffi2017icarl, mcdonnell2024ranpac, zhou2024expandable} construct and update prototypes, transforming inference into a matching between features and prototypes to reduce forgetting. Such approaches are straightforward, intuitive, and require only a small number of parameters, which have recently shown promising prospects. However, prototype-based methods cannot effectively classify samples with unsatisfied discriminative representations. We further analyze this limitation of prototype-based methods, discuss two key factors that contribute to this issue, and argue that this ultimately leads to a class-wise confusion problem.

Firstly, real data often resides on a manifold structure within latent spaces \citep{goodfellow2016deep, tenenbaum2000global, he2005face,lunga2013manifold}. As observed in Fig. \ref{fig:1-intro}(a), class 7 forms two distinct clusters in the latent feature space, while the prototype lies only in one of the clusters. This shows that a single discriminative vector has \textbf{limited representation capability} and fails to fit the manifold distribution, leading to the misclassification of the samples within the red box in Fig. \ref{fig:1-intro}(a) and resulting in class-wise confusion.
Secondly, since incremental learning cannot leverage the entire dataset, old exemplars often suffer from representational changes on the newly updated model, which is referred to as feature drift. Fig. \ref{fig:1-intro}(b) shows that old classes exhibit significant changes in class manifolds (class 3 shows a transformation of its manifold from the original spherical shape to a narrow elongated form in the new model) and feature dispersion that lead to \textbf{overlapped feature distributions} (samples within the red box), which confuses the model when distinguishing between these samples.

To address the aforementioned issues, this paper proposes a confusion-reduced auto-encoder classifier (CREATE) method as a solution. 
Firstly, considering that auto-encoders serve as manifold learners, learning a manifold structure for each class can enhance the stability of representations while effectively capturing the essential characteristics of the categories \citep{bengio2013representation, li2020rethink, zheng2022learning}, we utilize auto-encoder reconstructions to learn class distributions. Specifically, the auto-encoder module is applied for each class to capture low-dimensional essential structures and implicitly encode the feature distribution into it, thus tackling the problem of insufficient representational capacity. 
Secondly, we further design a confusion-aware separation loss that separates features of different classes in the class-specific latent space to learn a disentangling manifold for each class, thus mitigating the effect of overlapped features on the dynamically changing model.

The proposed method has the following advantages:
(1) The proposed auto-encoder reconstruction modules are representation condensed and lightweight. It can effectively fit the continuously changing manifolds of data.
(2) It can effectively discriminate samples that suffer from distribution overlap in the feature space, thereby reducing class confusion and forgetting.
Fig. \ref{fig:1-intro}(c) exhibits the manifold distributions in latent space.
Although features inevitably drift, our method can fit the manifolds well and eliminate representation overlap in the latent space by learning disentangled manifolds, thus reducing confusion.
The main contributions of this paper are summarized as follows:

$\bullet$ 
We identify the issue of class-wise confusion in incremental learning and propose a confusion-reduced auto-encoder classifier, which uses a lightweight auto-encoder for each class to learn a compact manifold. This paradigm exhibits a more expressive capability and can effectively adapt to feature drift at the reconstruction level.

$\bullet$ To further reduce the confusion of drifted features, we employ a confusion-aware separation loss at the class subspace level by disentangling samples from the distributions of other classes in the subspace.

$\bullet$ Our proposed method reduces class-wise confusion and has been validated through extensive experiments. It achieves better performance than the state-of-the-art methods up to $5.41\%$.

\section{Related Work}
\label{sec:relatedwork}

%-------------------------------------------------------------------------
\subsection{Class Incremental Learning}

Class incremental learning generally assumes that only a small number of samples can be stored for old classes, and task-id is not available in the inference phase. Existing methods can be divided into three main categories.

Knowledge retention-based methods aim to maintain the structure of old knowledge within the model and reduce knowledge variations to mitigate catastrophic forgetting. MGRB \citep{chen2022multi} constructs a knowledge structure for existing classes and is utilized for regularization when learning new classes. EDG \citep{gao2023exploring} maintains the global and local geometric structures of data in the mixed curvature space. DSGD \citep{fan2024dynamic} proposes a dynamic graph construction and preserves the invariance of the subgraph structure, which maintains instance associations during the CIL process.

Model expansion-based methods dynamically adjust the model architecture to adapt new classes. For example, DER \citep{yan2021der} expands a new backbone for each new task. The enhanced features from multiple backbone networks are concatenated for classification. FOSTER \citep{wang2022foster} adds an extra backbone to discover complementary features and eliminates redundant parameters by distillation. Memo \citep{zhou2023model} expands specialized blocks for new tasks to obtain diverse feature representations. 

Prototype-based methods establish a prototype for each task and update the prototypes in subsequent phases, classifying samples into the category of the most similar prototype in the inference phase. Some prototype-based methods use non-parametric class means as their prototypes. For example, iCaRL \citep{rebuffi2017icarl} suggests the nearest-mean-of-exemplars (NME) classifier determines the predicted label based on the distance from the sample features to the class center. SDC \citep{yu2020semantic} employs a metric loss-based embedding network and applies semantic drift compensation to adjust the prototypes closer to their correct positions. FCS \citep{Li2024FCS} uses one prototype to represent a class and adapts historical prototypes to the feature space on the new model by a feature calibration network. 
In recent years, parametric class prototypes have gained widespread use and achieved impressive performance. PODNet \citep{douillard2020podnet} learns multiple proxy vectors and predicts based on the local similarity classifier. RandPAC \citep{mcdonnell2024ranpac} proposes projecting features to an expanded dimension, which enhances the linear separability of prototypes. SEED \citep{Grzegorz2024divide} employs one Gaussian distribution for each class and performs an ensemble of Bayes classifiers.

\subsection{Class-wise Representation}

Many methods use prototypes to represent a class for classification. For example, \citet{snell2017prototypical} formulated prototypical networks for few-shot classification. It produces a distribution for query points using a softmax over distances to the prototypes in the embedding space. 
\citet{zhu2022multi} optimized the distances between query samples and different class prototypes to maintain boundaries at the class-granularity level.
\citet{Shi2023prototype} utilized expanded prototypes to express classes by self-supervised prototype reminiscence. 
\citet{huang2022learning} suggested representing a class with a prototype and multiple sub-prototypes, allowing the model to better capture the diversity within the same class. 
\citet{zhou2024prototype} proposed utilizing the centers of sub-clusters as a set of prototypes that comprehensively represent the characteristic properties. Apart from prototypes, recent methods utilize distributions to represent a category.
SEED \citep{Grzegorz2024divide} uses multivariate Gaussian distributions to represent each class and employs Bayesian classification from all experts. This method allows for more flexible and comprehensive class representations.
Considering the complex underlying structure of data distributions, \citet{lin2024class} modeled each class with a mixture of von Mises-Fisher distributions by multiple prototypes.

Auto-encoder structure is a type of manifold learner that can embed high-dimensional data into a low-dimensional manifold through nonlinear mapping. It is widely used for various tasks, such as anomaly detection, novel class detection, and few-shot learning. \citet{Kodirov_2017_CVPR} introduced a semantic auto-encoder that maps visual features to a low-dimensional semantic space, where incorporating class distribution information.
\citet{kim2019variational} utilized a variational auto-encoder to learn a latent space with strong generalization capabilities for unseen classes through data-prototype image pairs. After training, features are closely distributed around their corresponding prototype feature points in the latent space.
Our method considers utilizing auto-encoders to enhance representation capability, as data concentrates around a low-dimensional manifold in the latent space, which is a superior characteristic for CIL to learn efficient representations of classes.

\section{Methods}
\label{sec:methods}

\begin{figure*}[t]
\begin{center}
\includegraphics[width=1.0\linewidth]{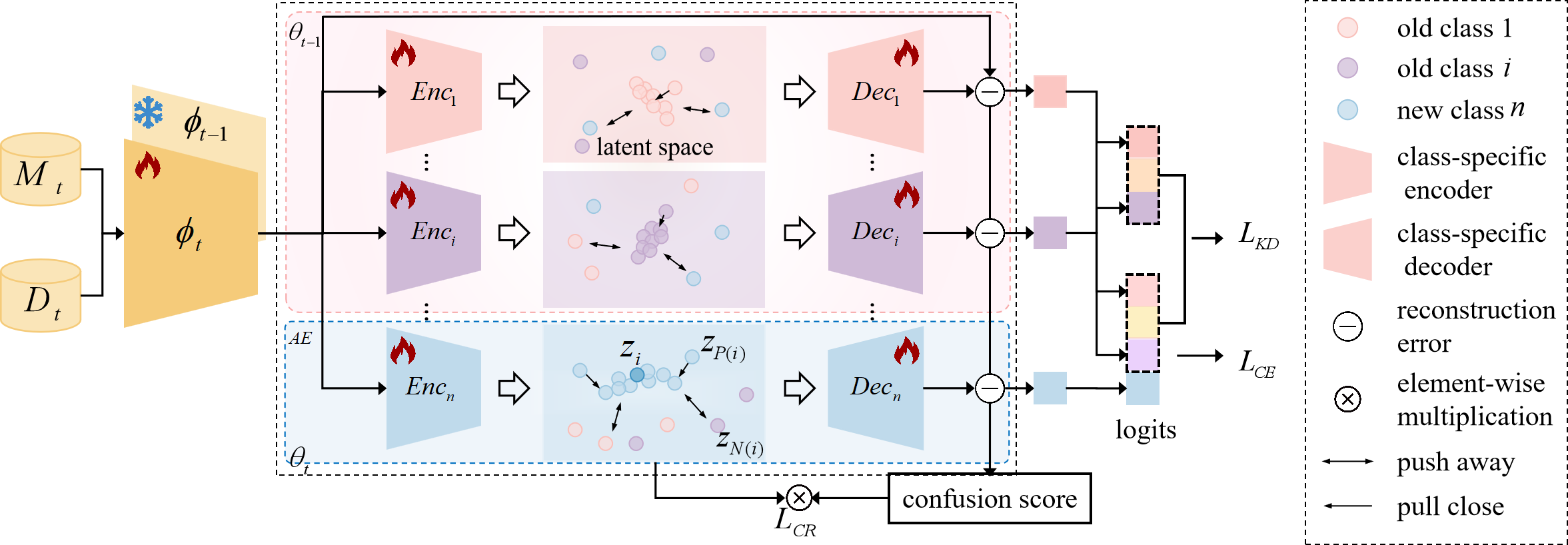}
\end{center}
\caption{Overview of the proposed confusion-reduced auto-encoder classifier (CREATE). The auto-encoder (AE) learns a subspace for each class and generates a latent class distribution. Preserving the trained old AEs facilitates the retention of knowledge from old classes, while making these AEs trainable ensures adaptability to updates in the feature extractor $\phi_t$.  
To further alleviate class confusion in class incremental learning, we employ a confusion-aware separation loss $L_{CR}$ to separate samples from other classes within each subspace.
}
\label{fig:4.1-architecture}
\end{figure*}

In this section, we give a description of the proposed confusion-reduced auto-encoder classifier. The problem setup is introduced in \cref{CIL_setup}. The core idea of the proposed method is to construct a lightweight learnable auto-encoder (AE) module for each class. Preserving these trained class-wise AEs can alleviate catastrophic forgetting since they represent accurate and complete class distributions, as detailed in \cref{class-specific-auto-encoders}.  
To further mitigate the accumulated class confusion of AEs arising from distribution drift in CIL, we also propose a confusion-aware latent space separation loss in \cref{Confusion-aware}. The framework is shown in \cref{fig:4.1-architecture}.

\subsection{Problem Setup}
\label{CIL_setup}
In CIL, we usually assume knowledge is not learned at once but from a sequence of $T$ tasks (phases). $\mathcal{D}_t=\{(x_i^t, y_i^t)\}_{i=1}^{n_t}$ represents $n_t$ samples from the task $t$. $C_t$ is the number of classes seen by phase $t$. In rehearsal-based methods, $\mathcal{M}_t$ represents the memory buffer in the $t$-th task. Therefore, the training dataset for task $t$ is $\mathcal{D}_t \bigcup \mathcal{M}_t$. Note that the sets of new classes learned in different incremental tasks are mutually exclusive. The model in phase $t$ can be decomposed into feature extractor $\phi_t$ and classification module $\theta_t$.

\subsection{Learning Disentangled Manifolds by the Auto-encoder Classifier}
\label{class-specific-auto-encoders}

To effectively depict class distributions and also avoid excessive computation, we consider constructing a lightweight auto-encoder module for each category so that the original features can be mapped into the corresponding subspace. The learned subspaces compress the patterns of samples into a compact and continuous low-dimensional manifold, allowing the auto-encoder to better model the distribution of the class. The manifolds of each class reside in their respective subspaces, remaining unaffected by others, and thus are disentangled.
To adapt to new tasks and remember old classes, the previous auto-encoders are retained and kept trainable, and class-specific auto-encoders for new classes are appended. 
We hope the class-specific auto-encoders can identify samples of their own categories within the dataset. Therefore, we consider using reconstruction error to measure the degree of consistency between the samples and the auto-encoder subspaces.

The overall framework is shown in \cref{fig:4.1-architecture}. The feature extractor $\phi$ is utilized to obtain features, followed by a group of class-specific auto-encoders that compress and reconstruct these features for classification.
Specifically, for class $i$, we construct an auto-encoder $AE_i$, consisting of an encoder $f_i: \mathbb{R}^{d} \to \mathbb{R}^l (d > l)$ learns a mapping that projects the original features into latent subspace, and a decoder $g_i: \mathbb{R}^{l} \to \mathbb{R}^d$ (symmetric in structure to encoder) that reconstructs features based on the latent representations. Both encoders and decoders use a 1$\times$1 convolutional layer with a $\tanh$ activation function. The output channel size $l$ for $f_i$ and the input channel size for $g_i$ are both set to 32. It takes features $h=\phi(x)$ as input (here, the task index $t$ is ignored for clarity), and outputs the reconstructed embeddings on each module: $\widetilde{h}_i=g_i(f_i(h))$, where $i=1,...,C_t$. The reconstruction error of representation $h$ on the $i$-th auto-encoder is noted as follows:
\begin{equation}
    \epsilon_i = \parallel \widetilde{h}_i - h \parallel .
\end{equation}

We use reconstruction errors as the classification metric. On the one hand, we hope that samples on their corresponding auto-encoders have the smallest reconstruction error, which indicates that the semantic knowledge and the class-specific manifold are effectively captured and learned. This goal can be achieved by minimizing the reconstruction errors to zero for samples on their ground-truth auto-encoders. On the other hand, we expect samples processed by modules that do not belong to their specific classes to exhibit larger errors, indicating that mapping samples to the wrong subspaces results in significantly mismatches. Therefore, we process the reconstruction errors as \cref{eq:prob} to obtain the predicted probability of the sample $x$:
\begin{equation}
    p_i = \frac{\exp (-\alpha \epsilon_i)}{\sum_{j=1}^{C_t}{\exp (-\alpha \epsilon_j)}},
\label{eq:prob}
\end{equation}
where $\alpha$ is a positive hyper-parameter used to adjust the scale of the reconstruction errors. We can see that the probability $p_i$ is negatively correlated with the distance between the reconstructed feature on $i$-th auto-encoder and the original feature. Then, we employ the cross-entropy loss function to measure the difference between the predicted probability distribution and the ground truth, which is helpful for inter-class discrimination:
\begin{equation}
    L_{CE}=-\sum_{i=1}^{C_t}  y_i \log p_i.
\label{eq:lossce}
\end{equation}

When a new task arrives, auto-encoders of new classes are added, and auto-encoders of old classes are reserved and updated to new distributions.
To mitigate forgetting of old classes, we apply the distillation loss on the logits-level, formulated as:
\begin{equation}
    L_{KD}= -\sum_{i=1}^{C_{t-1}} \frac{\exp(-\frac{\alpha {\epsilon_i}^{\prime}}{\tau_{d}})}{\sum_{j=1}^{C_{t-1}}{\exp (- \frac{\alpha {\epsilon_j}^{\prime} }{\tau_{d}})}} \log \frac{\exp(-\frac{\alpha \epsilon_i}{\tau_{d}}  )}{\sum_{j=1}^{C_{t-1}}{\exp (-\frac{\alpha \epsilon_j}{\tau_{d}})}},
\label{eq:losskd}
\end{equation}

where ${\epsilon_i}^{\prime}$ is the error provided by the old network. $\tau_{d}$ is the distillation temperature. 
Note that the feature extractor and auto-encoders remain unfrozen during training to adapt to new tasks. This is because the feature extractor continuously updates to accommodate the learning of new tasks. The representations of old class samples also change on the newly updated model. If the auto-encoder for an old class is frozen, its subspace will remain unchanged, making it unable to adapt to and match the shifted features of old classes, which may lead to a decrease in stability. 

\subsection{Confusion-aware Latent Space Separation}
\label{Confusion-aware}
Due to the feature shift in incremental learning, confusion between classes is becoming increasingly severe, leading to catastrophic forgetting. 
We propose further class separation within the subspace to reduce confusion when distinguishing overlapped features. 
This process is essential, as features of different classes may be distributed in similar positions after a shift. Therefore, compressed features become relatively similar to each other on the manifold, and so do the reconstructed features. The similar reconstruction errors thus confuse classification.

Firstly, we measure the confusion score $s_i$ for sample $i$ expressed as follows: 

\begin{equation}
    s_{i}=\frac{|{\epsilon_i}_{(2)} -
 {\epsilon_i}_{(1)}|}{{\epsilon_i}_{\max} - {\epsilon_i}_{(1)}},
\label{eq:score}
\end{equation}
where ${\epsilon_i}_{(1)}$ and ${\epsilon_i}_{(2)}$ are the smallest and the second smallest values in an error sequence $\epsilon_i$, respectively. A smaller $s$ for a sample suggests more similar reconstruction errors between two auto-encoders, indicating significant confusion.

Therefore, we suggest that samples should be located far from the manifold region in their non-ground truth auto-encoders' latent space. We employ a contrastive loss in the class-specific subspaces, written as follows:

\begin{equation}
   L_{CST} = \sum_{i=1}^{C_t} - \frac{1}{| P(i) |}\sum_{p\in P(i)} \log\frac{\exp(z_i^T \cdot z_{p_i} / \tau_{r})}{\sum_{k\in S(i)} \exp(z_i^T \cdot z_{k_i} / \tau_{r})},
\label{eq:cont}
\end{equation}
where $z_i=f_i(\phi(x))$, $\tau_{r}$ is temperature. $P(i)$ represents the positive samples set of class $i$, and $z_{p_i}$ is the latent representation on the $i$-th class auto-encoder for samples that share the same label as $z_i$. $N(i)$ denotes the set of negative samples of class $i$. $S(i)=P(i)\cup N(i)$.

Optimizing \cref{eq:cont} allows us to learn more accurate manifold and reduce class-wise confusion problem. This is achieved by maximizing the mutual information between positive samples, which draws positive pairs closer in the latent space and fits a low-dimensional manifold, while simultaneously minimizing the mutual information between negative samples, thereby pushing them further away from the manifold region of the particular class.

Considering the varying degrees of sample confusion, we convert the confusion scores into weights by \cref{eq:weight} and obtain a confusion-reduce contrastive loss function formulated as \cref{eq:wcont}.

\begin{equation}
    w_{i}=1+\exp(-\beta s_i),
\label{eq:weight}
\end{equation}

\begin{equation}
    L_{CR}=\sum_{i=1}^{C_t} \frac{-1}{| P(i) |}\sum_{p\in P(i)} w_{i}\log\frac{\exp(z_i^T \cdot z_{p_i} / \tau_{r})}{\sum_{k\in S(i)} \exp(z_i^T \cdot z_{k_i} / \tau_{r})}.
\label{eq:wcont}
\end{equation}

The complete loss for this model is formulated as:
\begin{equation}
    L = L_{CE}+L_{KD}+\lambda L_{CR},
\label{eq:total}
\end{equation}
where $\lambda$ is a weighting parameter that balances feature and latent space separation. The supplementary material provides a detailed description of the algorithm.

In summary, our proposed model addresses two issues in existing prototype-based methods: insufficient representational capacity and sensitivity to overlapped features. We enhance representational capability by utilizing a group of auto-encoders to capture the disentangled manifold of each class and improve class separation in the latent representation to reduce sensitivity to overlapped features. Ultimately, this approach helps mitigate the confusion problem encountered in incremental learning.

\section{Experiments}
\label{Sec4}

\begin{table*}[t]
\caption{Last and average accuracy of different methods on CIFAR100. The best performance is highlighted in bold, while the second-best performance is indicated with underline.}
% \small
\renewcommand\tabcolsep{4.5pt}     
\renewcommand\arraystretch{1}  
\begin{center}
\begin{tabular}{ccccccccc}
\toprule
\multirow{3}{*}{Methods} & \multicolumn{4}{c}{CIFAR100 B0}                                           & \multicolumn{4}{c}{CIFAR100 B50}                                                  \\
\cmidrule(lr){2-5} \cmidrule(lr){6-9}
& \multicolumn{2}{c}{Inc 10}             & \multicolumn{2}{c}{Inc 20}              & \multicolumn{2}{c}{Inc 5}               & \multicolumn{2}{c}{Inc 10}              \\
\cmidrule(lr){2-3} \cmidrule(lr){4-5} \cmidrule(lr){6-7} \cmidrule(lr){8-9}
& Last           & Avg               & Last           & Avg              & Last           & Avg               & Last           & Avg            \\ 
\midrule
iCaRL (CVPR' 2017)                    & 49.52          & 64.42           & 54.23          & 67.00           & 47.27          & 53.21           & 52.04          & 61.29          \\
PODNet (ECCV' 2020)                    & 36.78          & 55.22           & 49.08          & 62.96           & 52.11          & 62.38           & 55.21          & 64.45          \\
WA (CVPR' 2020)                         & 52.30          & 67.09           & 57.97          & 68.51           & 48.01          & 55.90            & 55.85          & 64.32          \\
DER (CVPR' 2021)                        & 60.96          & 71.25           & 62.40          & 70.82          & 56.57          & 64.50            & 61.94          & 68.24          \\
Foster (ECCV' 2022)                      & 62.54          & 72.81          & 64.55          & 72.54           & 60.44          & 67.94           & 64.01          & 70.10          \\
DyTox (CVPR' 2022)                    & 58.72          & 71.07      & 64.22          & \underline{73.05}    & -              & -              & 60.35          & 69.07          \\ 
BEEF (ICLR' 2023)                        & 60.98          & 71.94           & 62.58          & 72.31          & 63.51          & \underline{70.71}    & 65.24          & \underline{71.70}    \\
DGR (CVPR' 2024)                        & 57.10          & 68.40           & 61.90          & 70.70            & 54.70          & 61.90          & 58.90          & 66.50          \\
DSGD (AAAI' 2024)                        & \underline{63.18} & \underline{73.01}    & \underline{67.67}    & 72.91           & \underline{ 63.53}    & 68.14          & \underline{65.83}    & 70.02          \\
\midrule
CREATE                    & \textbf{63.69} & \textbf{75.60}  & \textbf{69.99} & \textbf{78.46}   & \textbf{63.58} & \textbf{72.27}  & \textbf{68.40} & \textbf{75.52} \\
Gain ($\Delta$)                     & \color{blue}{+0.51}             & \color{blue}{+2.59}            & \color{blue}{+2.32}       & \color{blue}{+5.41}             & \color{blue}{+0.05}          & \color{blue}{+1.56}         & \color{blue}{+2.57}         & \color{blue}{+3.82}         
\\ 
\bottomrule

\end{tabular}
\label{tab:c100}
\end{center}
\end{table*}

\begin{table*}[t]
\caption{Last and average accuracy of different methods on ImageNet100. The best performance is highlighted in bold, while the second-best performance is indicated with underline. ``\#P'' represents the number of parameters (million).}
% \small
\renewcommand\tabcolsep{4.0pt}     
\renewcommand\arraystretch{1}  
\begin{center}
\begin{tabular}{ccccccccccccc}
\toprule
\multirow{3}{*}{Methods} & \multicolumn{6}{c}{ImageNet100 B0}                                          & \multicolumn{6}{c}{ImageNet100 B50}                                                                          \\
\cmidrule(lr){2-7} \cmidrule(lr){8-13}
& \multicolumn{3}{c}{Inc 10}        & \multicolumn{3}{c}{Inc 20}              & \multicolumn{3}{c}{Inc 5}                           & \multicolumn{3}{c}{Inc 10}                             \\
\cmidrule(lr){2-4} \cmidrule(lr){5-7} \cmidrule(lr){8-10} \cmidrule(lr){11-13}
& \#P   & Last        & Avg         & \#P   & Last           & Avg            & \#P                  & Last        & Avg            & \#P                  & Last           & Avg            \\ \midrule
DyTox                    & 11.00 & 61.78       & 73.40       & 11.00 & 68.78          & 76.81          & 11.00                & -           & -              & 11.00                & 65.76          & 74.65          \\
iCaRL                    & 11.17 & 50.98       & 67.11       & 11.17 & 61.50          & 73.57          & 11.17                & 50.52       & 57.92          & 11.17                & 53.68          & 62.56          \\
PODNet                   & 11.17 & 45.40       & 64.03       & 11.17 & 58.04          & 71.99          & 11.17                & 64.70       & 72.59          & 11.17                & 62.94          & 73.83          \\
WA                       & 11.17 & 55.04       & 68.60       & 11.17 & 64.84          & 74.44          & 11.17                & 50.16           & 61.61              & 11.17                & 56.64          & 65.81          \\
Foster                   & 11.17 & 60.58       & 69.36       & 11.17 & 68.88          & 75.27          & 11.17                & 67.78     &      76.21          & 11.17                & 63.12          & 69.85          \\
DGR                      & 11.17 & 64.00       & 72.80       & 11.17 & 71.10          & 77.50          & 11.17                & 62.60       & 70.50          & 11.17                & 69.30          & 74.90          \\
DER                      & 111.7 & 66.84       & 77.08       & 55.85 & \underline{72.10}    & \underline{78.56}    & 122.87               &      69.30       &   \underline{77.69}             & 67.02                & 71.10          & 77.57          \\
DSGD                     & 111.7 & \textbf{68.32}       & 75.68       & 55.85 & 71.76          & 77.07          & 122.87               & \underline{69.50} & 77.20    & 67.02                & 73.01          & 80.30          \\
BEEF                     & 111.7 & 66.59 & \textbf{78.41} & 55.85 & -              & -              & 122.87               & -           & -              & 67.02                & \underline{74.62}    & \underline{80.52}    \\ \midrule
CREATE                   & 14.44 & \underline{66.94}       & \underline{77.95}       & 14.44 & \textbf{74.34} & \textbf{81.37} & 14.44                & \textbf{71.42}       & \textbf{79.44} & 14.44                & \textbf{77.06} & \textbf{82.43} \\
Gain ($\Delta$)                 &       &          &             &       & \color{blue}{+2.24}          &\color{blue}{ +2.81}         & \multicolumn{1}{l}{} &\color{blue}{ +1.92}       & \color{blue}{+1.75}          & \multicolumn{1}{l}{} & \color{blue}{+2.44}          & \color{blue}{+1.91}
\\ \bottomrule

\end{tabular}
\label{tab:i00}
\end{center}
\end{table*}

\begin{table*}[!ht]
\centering
\caption{Ablations study in our method. We report the accuracy of each phase and the average accuracy under CIFAR100 Base50 Inc10.}
\renewcommand\tabcolsep{4.5pt}     
\renewcommand\arraystretch{1}  
\scalebox{1.0}{
\begin{tabular}{ccccccccccccc}
\hline
Comp.    & NME & AEs & $L_{CR}$ & 50    & 60    & 70    & 80    & 90    & 100   & Avg            \\ \hline
NME & \checkmark   &     &      &    
84.80 &	75.90 &	69.94 &	65.82 &	63.03 &	62.09 &	70.26 
\\
Ours-AE  &     & \checkmark   &          & 84.20 &	79.75 &	76.13 &	71.26 &	68.47 &	65.31 &	74.19 
        \\
Ours     &     & \checkmark   & \checkmark        & 84.64 &	80.42 &	76.79 &	72.80 &	70.09 &	68.40 	 
 & \textbf{75.52} 
\\
\hline
\end{tabular}
}
\label{tab:ablation}
\end{table*}

\subsection{Experimental Setups}

\textbf{Datasets.} CIFAR100 \citep{krizhevsky2009learning} consists of 32x32 pixel images and has 100 classes. Each class contains 600 images, with 500 for training and 100 for testing. ImageNet100 \citep{deng2009imagenet} is selected from the ImageNet-1000 dataset, comprising 100 distinct classes. Each class contains about 1300 images for training and 500 images for testing.

\textbf{Protocols.} For CIFAR100 and ImageNet100, we evaluate the proposed method on two widely used protocols: Base0 for learning from scratch and Base50 for learning from half. In Base0, classes are evenly divided. Inc10 and Inc20 refer to tasks containing 10 and 20 classes, incrementally learning until all classes are covered. Up to 2,000 exemplars can be stored. Base50 refers to a model that learns 50 classes in the first phase, and then learns the remaining 50 classes in Inc5 mode (5 classes per task) or Inc10 mode (10 classes per task). The memory buffer is set to 20 exemplars per class. We denote the accuracy after task $t$ as $A_t$ and use the final phase accuracy $A_T$ and average incremental accuracy $\bar A = \frac{1}{T}\sum_{t=1}^T A_t$ for comparison. We use ``\#P'' to denote the parameters count in million after the final phase.

\textbf{Implementation details.} The proposed method is implemented with PyTorch and PyCIL \citep{zhou2023pycil}. Experiments are run on the NVIDIA RTX3090 GPU. We employ ResNet18 trained from scratch as the feature extractor for both CIFAR100 and ImageNet100. We adopt an SGD optimizer with a weight decay of 2e-4 and a momentum of 0.9. We train the model for 200 epochs in the initial phase and 120 epochs in the subsequent incremental phase. The batch size is 128, and the initial learning rate is 0.1. We set the hyper-parameter $\alpha$ to 0.1, $\beta$ to 2 and $\lambda$ to 1. The temperature $\tau_{d}$ in $L_{KD}$ is set to 2, and temperature $\tau_{r}$ in $L_{CR}$ is set to 0.1 for all experiments. 

\subsection{Experimental Results}
\textbf{Comparative performance.}
\Cref{tab:c100} and \Cref{tab:i00} presents the comparative experiments on CIFAR100 and ImageNet100.
We run four settings on the CIFAR100 and Imagenet100 dataset including both learning from scratch and learning from half, incremental learning 5 and 10 phases, and report the last phase accuracy and the average accuracy.  It can be seen that our method surpasses the best results in both last accuracy and average incremental accuracy by $1.36\%$ and $3.35\%$, respectively, on average across the four settings on CIFAR100. The greater enhancements in average incremental accuracy indicate that our method ensures a steady improvement throughout the entire learning progress, rather than only in the final phase. 
In the learning from half setting Base50 Inc10, our method gets an average accuracy improvement of $3.82\%$ over BEEF \citep{wang2023beef} and a final accuracy enhancement of $2.57\%$ over DSGD \citep{fan2024dynamic}.
In the learning from scratch setting Base0 Inc20, our method outperforms DSGD by $2.32\%$ in the last accuracy and exceeds DyTox by $5.41\%$ in the average accuracy. 

\Cref{tab:i00} presents the comparative experiments on  ImageNet100. Our method also surpasses existing state-of-the-art methods by a significant margin in most settings.
In Base50 Inc10, our method gets an average accuracy improvement of $1.91\%$ and a final accuracy enhancement of $2.44\%$ over BEEF. As for Base0 Inc10, our method achieves the second-highest accuracy, while it has only about one-eighth the number of parameters compared with DSGD and BEEF.
Our framework can capture intrinsic manifold distribution and improve the discriminative ability of confusing features, yielding remarkable results in both Base0 and Base50 protocols.

\textbf{Parameter efficiency.}
In addition to comparative accuracy, our method exhibits parameter efficiency in ImageNet100, as presented in \cref{tab:i00}. Our proposed method consistently outperforms the state-of-the-art methods by $1.75\%$ and $1.91\%$ in the average accuracy of learning from half settings, and reduces the parameter by $88.2\%$ and $78.5\%$, respectively. 
Although BEEF achieves a higher average accuracy in the Base10 Inc10 setting, it comes at the cost of approximately ten times the parameter scale compared to our method.
Employing an auto-encoder architecture to learn the disentangled class manifolds can achieve more competitive confusion reduction results and efficient outcome with a significantly fewer parameter costs compared to other separation methods.

\begin{figure}[t]
\centering
\begin{subfigure}{0.49\linewidth}
\includegraphics[width=\linewidth]{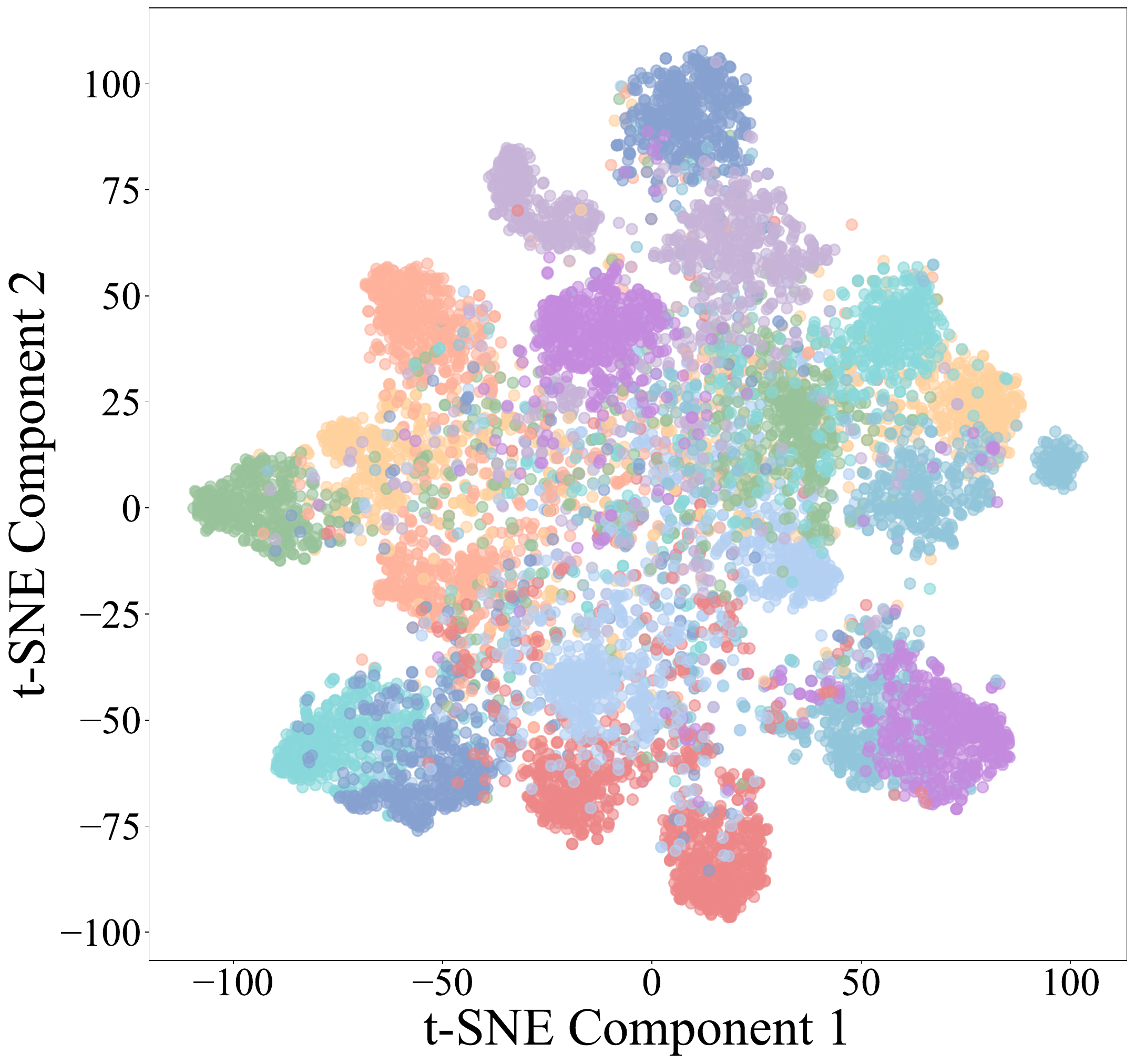}
% \fbox{\rule{0pt}{2in} \rule{.9\linewidth}{0pt}}
\caption{WA}
\end{subfigure}
\begin{subfigure}{0.49\linewidth}
\includegraphics[width=\linewidth]{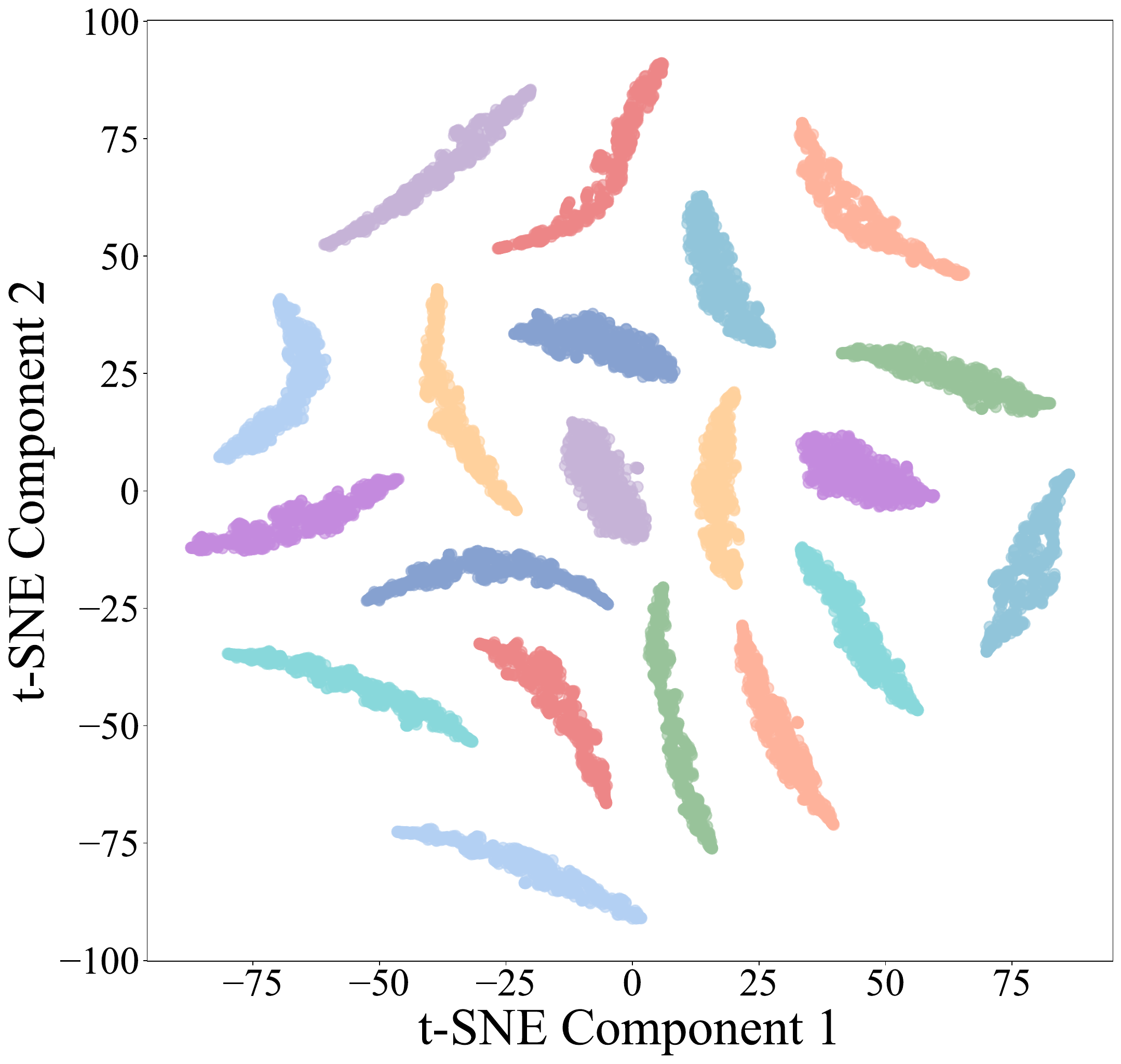}
\caption{CREATE}
\end{subfigure}
\caption{Visualization of feature distributions on WA and Ours under CIFAR100 Base50 Inc5 final phase. CREATE can well distinguish class distributions.}
\label{fig:4.3-ablation-tsne}
\end{figure}

\textbf{Visualization of feature distributions.}
In \cref{fig:4.3-ablation-tsne}(a), we plot the distributions of 20 classes (including both old and new classes) in the feature space of WA for clearer observation. The feature distribution of the compared method still exhibits overlap. 
In contrast, \cref{fig:4.3-ablation-tsne}(b) shows the t-SNE of the same class distributions in the class-specific latent space of our proposed method (CREATE). It highlights the superiority of the class-specific auto-encoder which can learn a more separated manifold distribution for each class and reduce the overlap in the subspace, thus alleviating class confusion for CIL.

\subsection{Ablation Study}

In this section, we conduct experiments to verify the effectiveness of the components. We validated the following four aspects: (1) quantitative analysis on component effectiveness, (2) class confusion reduction analysis, (3) impact of structure of class-specific auto-encoder, and (4) impact of hyper-parameters.

\begin{figure}[t]
\centering
\begin{subfigure}{0.49\linewidth}
\includegraphics[width=\linewidth]{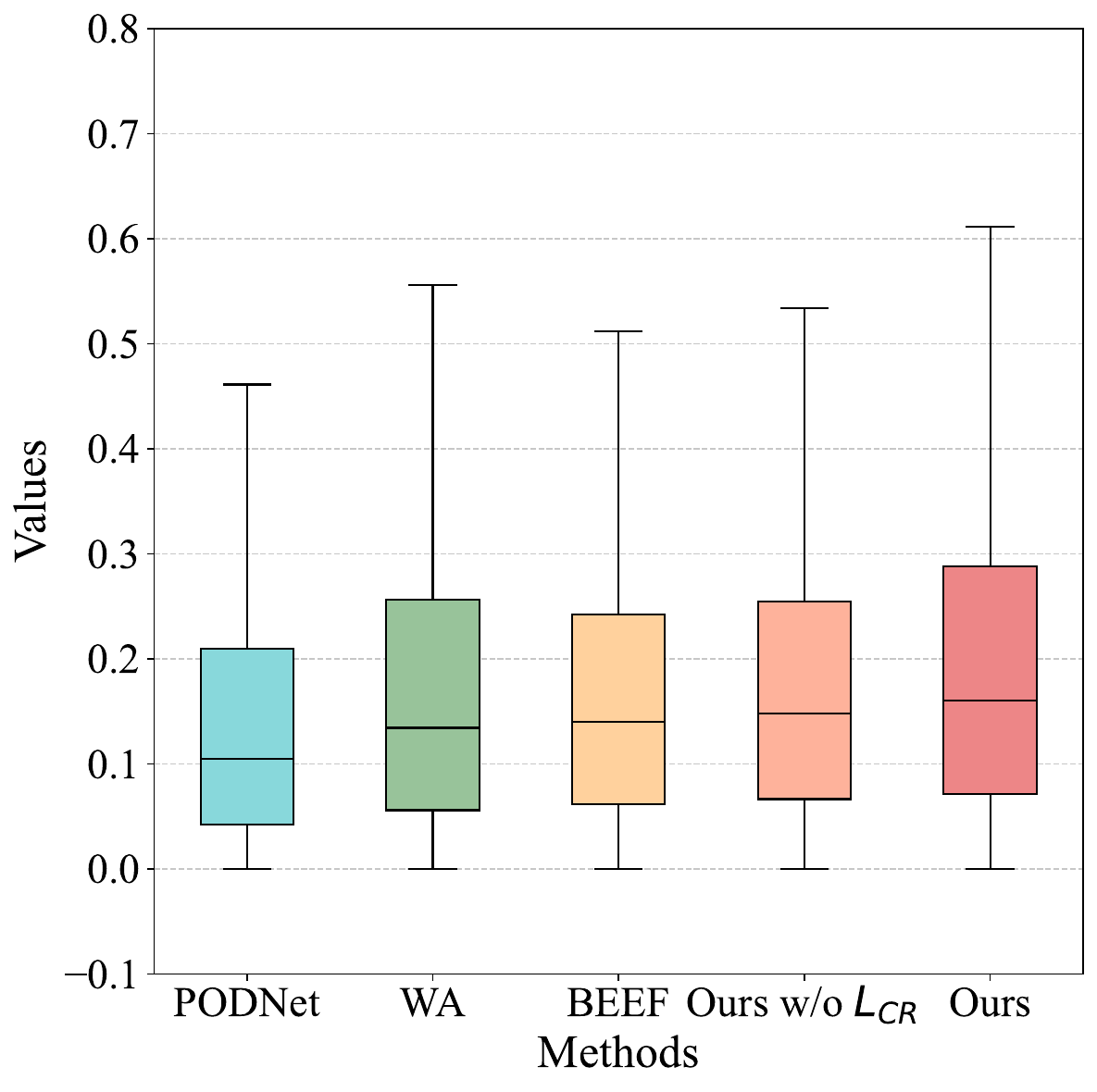}
\caption{Confusion degree}
\end{subfigure}
\begin{subfigure}{0.49\linewidth}
\includegraphics[width=\linewidth]{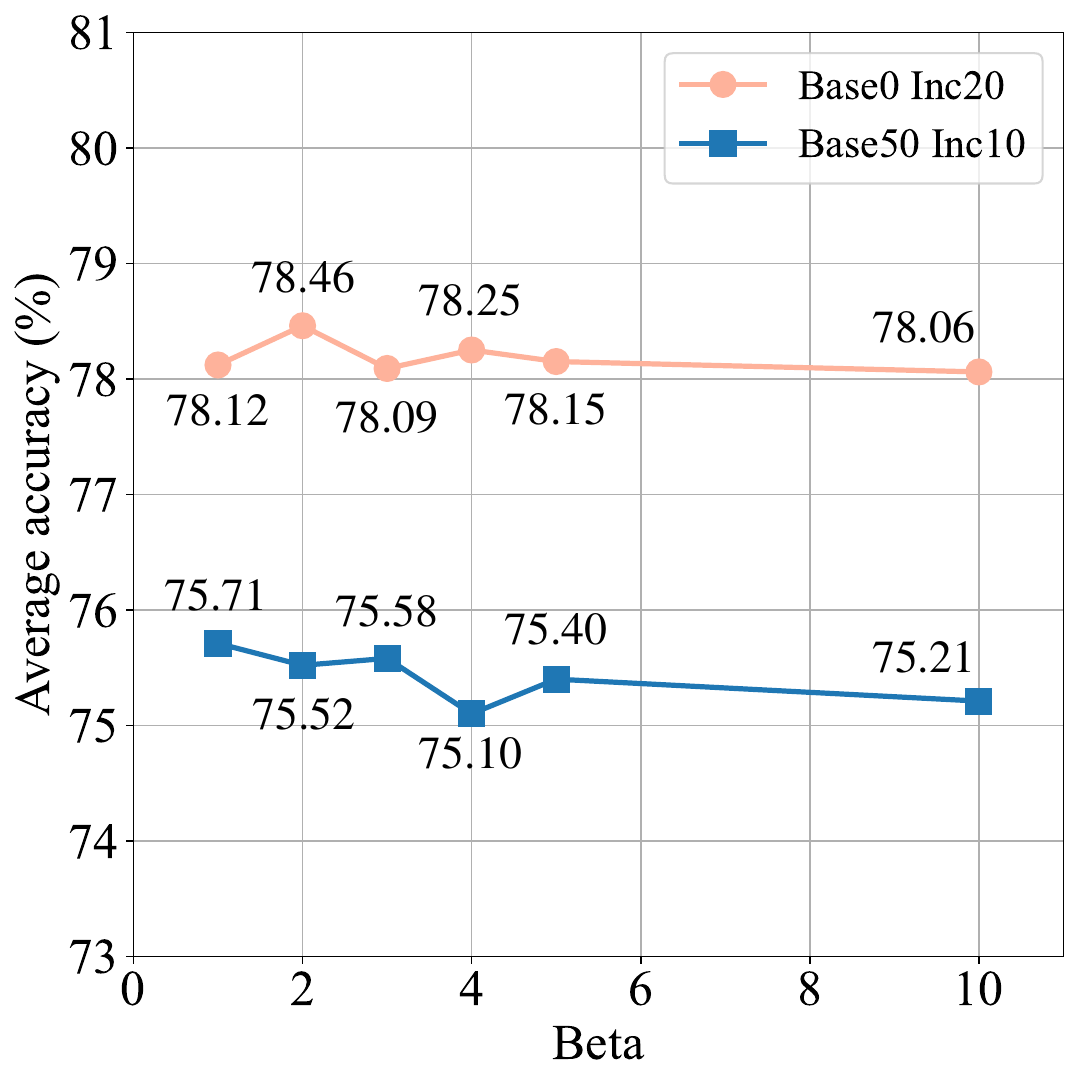}
\caption{Hyper-parameter $\beta$ sensitivity}
\end{subfigure}
\caption{Analysis of confusion degree and hyper-parameter sensitivity.}
\label{fig:4.3-ablation-confusion}
\end{figure}

\begin{figure}[t]
\centering
\begin{subfigure}{0.49\linewidth}
\includegraphics[width=\linewidth]{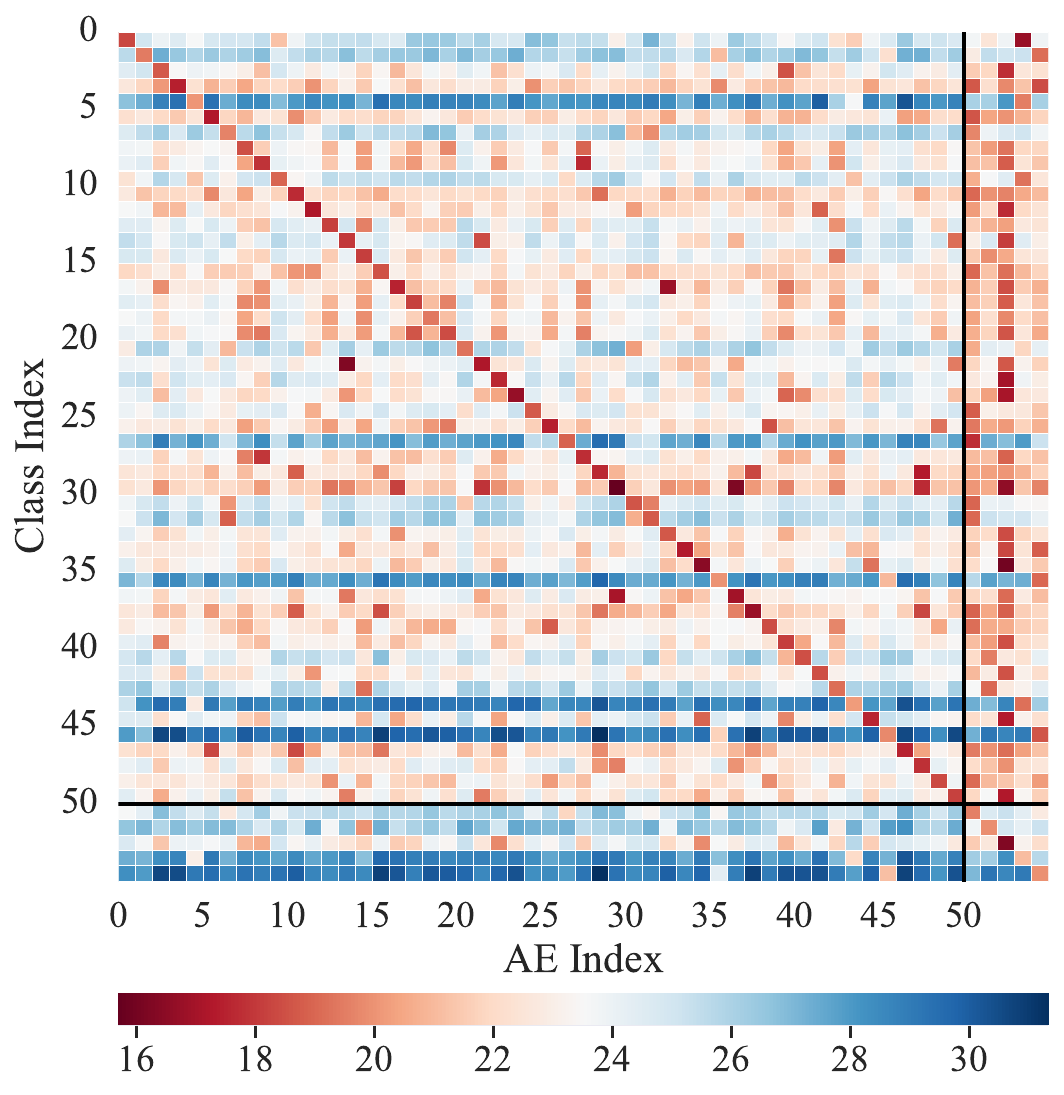}
\caption{Ours w/o $L_{CR}$}
\end{subfigure}
\begin{subfigure}{0.49\linewidth}
\includegraphics[width=\linewidth]{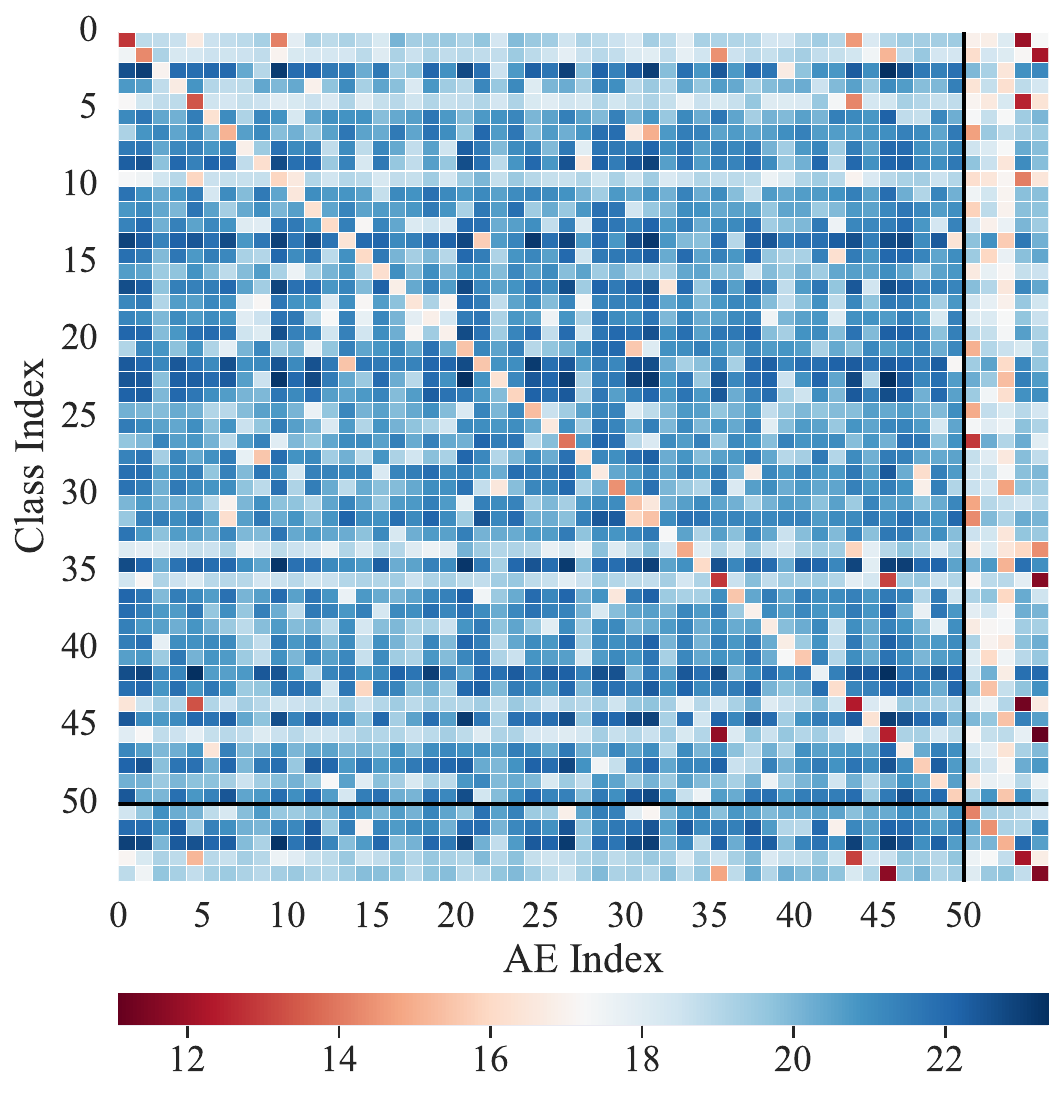}
\caption{Ours w/ $L_{CR}$}
\end{subfigure}
\caption{Reconstruction errors of misclassified data in CIFAR100 Base50 Inc5 phase2.}
\label{fig:4.2-confusion}
\end{figure}

\textbf{Effectiveness of components.} We conduct ablation experiments on CIFAR100 Base50 Inc10. \textbf{NME} means predicting the labels of test samples by a nearest-mean-of-exemplars classifier that calculates distances between their embeddings and prototypes of each class. Prototypes are updated on the new model after learning a new task. \textbf{Ours-AE} infers labels based on class-specific auto-encoders without $L_{CR}$.  \textbf{Ours} contains both the proposed framework and $L_{CR}$. As shown in \cref{tab:ablation}, the accuracy increases as we gradually add the proposed components. The proposed framework improves the average accuracy by $3.93\%$ over the classical prototype-based method. The final composition of the method raises the performance to $75.52\%$.

\textbf{Class confusion analysis.}
We draw the box plot of confusion scores defined by \cref{eq:score} for multiple methods on the test set in \cref{fig:4.3-ablation-confusion}(a), where for other methods, the variable $e$ is replaced with logits. A smaller confusion score signifies a greater degree of confusion in category predictions. 
Our method exhibits higher confusion scores compared to other approaches. 
It can be observed that both the mean value of the confusion score and the upper quartile are higher than those of the comparison methods.
This indicates that our model can effectively distinguish and reduce confusion when faced with shifted features.

\Cref{fig:4.2-confusion} shows the impact of our confusion-aware separation loss after learning the second phase. We find that implementing $L_{CR}$ increases the reconstruction error of the auto-encoder for samples that do not belong to their respective semantic categories. Therefore, $L_{CR}$ can enhance class separation in the latent space, thereby addressing the confusion caused by feature overlap in CIL.

\textbf{Architecture of class-specific auto-encoders.} 
We investigate the impact of the structure of the auto-encoder on learning disentangled manifolds. As shown in \cref{abla-AEchannel}, the first three columns indicate the use of a single convolutional layer, while the last two columns represent the use of two convolutional layers. We can observe that a single convolutional layer achieves the highest performance. This may be because simple architecture is sufficient to learn disentangled manifolds. Furthermore, there is a positive correlation between the increase in latent space dimensions and model performance. This may be attributed to the ability of larger latent spaces to capture more information, particularly in the presence of complex data. However, the benefits of increasing the dimension are limited to 0.07$\%$ in average accuracy when increasing the dimension from 32 to 64.

\begin{table}[t]
\caption{Architecture analysis on CIFAR100 Base50 Inc10. ``Channel'' represents the channel number of the convolutional layer in each encoder. }
% \small
\renewcommand\tabcolsep{4.0pt}
\renewcommand\arraystretch{1}  
\begin{center}
\begin{tabular}{cccccc}
\hline
Channel & 16    & 32    & 64    & 32,16 & 64,32 \\ \hline
Last    & 67.59 & 68.40  & 68.49 & 64.68 & 66.13 \\
Avg     & 75.30  & 75.52 & 75.59 & 73.48 & 73.98 \\ \hline
\end{tabular}
\end{center}
\label{abla-AEchannel}
\end{table}

\textbf{Impact of hyper-parameters.} The hyper-parameters used in the method are $\alpha$ for scaling reconstruction errors and $\beta$ for controlling confusion weights of samples. We set $\alpha$ to 0.1 in all experiments. Thus, we conducted experiments on the remaining hyper-parameter $\beta$. As shown in \cref{fig:4.3-ablation-confusion}(b), for $\beta \in \{1, 2, 3, 4, 5, 10\}$, the average accuracy remains relatively stable in various settings.

\section{Conclusion}
\label{sec:conclusion}
In this paper, we propose an auto-encoder classifier to reduce class-wise confusion in incremental learning. It employs a lightweight auto-encoder module and learns disentangled manifolds for each class to represent their distribution. Moreover, it constrains latent spaces by a confusion-aware separation loss that enhances class separability. This approach addresses the problem of insufficient representational capacity and severe class confusion in the dynamic distribution-changing situation of prototype-based CIL methods. Experimental results show that our method achieves state-of-the-art performance in various scenarios. 

{
    \small
    \bibliographystyle{ieeenat_fullname}
    \bibliography{main}
}

% compile the supplementary material
% \input{sec/0_supp}
% \input{sec/X_suppl}
% {
%     \small
%     \bibliographystyle{ieeenat_fullname}
%     \bibliography{main}
% }

\end{document}